\pdfoutput=1

\documentclass[11pt]{article}

\usepackage[]{acl}

\usepackage{times}
\usepackage{latexsym}

\usepackage{times}
\usepackage{latexsym}
\usepackage{hyperref}
\usepackage{graphicx}
\usepackage{makecell}
\usepackage[subrefformat=parens]{subcaption}
\usepackage{caption}
\usepackage{subcaption}
\usepackage{booktabs}
\usepackage{multicol}
\usepackage{graphicx}
\usepackage{multirow}
\usepackage{microtype}
\usepackage{makecell}
\usepackage{pgfplots}

\usepackage[T1]{fontenc}

\usepackage[utf8]{inputenc}

\usepackage{microtype}

\usepackage{inconsolata}
\usepackage{amssymb}
\usepackage{amsmath}
\usepackage{multicol}
\usepackage{multirow}

\title{Enhancing Financial Sentiment Analysis with Expert-Designed Hint}

\author{Chung-Chi Chen,\textsuperscript{1} Hiroya Takamura,\textsuperscript{1} Ichiro Kobayashi,\textsuperscript{2}  Yusuke Miyao\textsuperscript{3}
\\
 \textsuperscript{1} Artificial Intelligence Research Center, AIST, Japan \\
\textsuperscript{2} Ochanomizu University, Japan \\
 \textsuperscript{3} University of Tokyo, Japan \\
   \texttt{c.c.chen@acm.org, takamura.hiroya@aist.go.jp,}\\ \texttt{koba@is.ocha.ac.jp, yusuke@is.s.u-tokyo.ac.jp}\\
}

\begin{document}
\maketitle
\begin{abstract}
This paper investigates the role of expert-designed hint in enhancing sentiment analysis on financial social media posts. We explore the capability of large language models (LLMs) to empathize with writer perspectives and analyze sentiments. Our findings reveal that expert-designed hint, i.e., pointing out the importance of numbers, significantly improve performances across various LLMs, particularly in cases requiring perspective-taking skills. Further analysis on tweets containing different types of numerical data demonstrates that the inclusion of expert-designed hint leads to notable improvements in sentiment analysis performance, especially for tweets with monetary-related numbers. Our findings contribute to the ongoing discussion on the applicability of Theory of Mind in NLP and open new avenues for improving sentiment analysis in financial domains through the strategic use of expert knowledge.
\end{abstract}

\section{Introduction} 
\label{sec:introduction}

Financial sentiment analysis, which classifies given text into bullish or bearish categories, has long been a topic of research in NLP~\cite{baker2007investor,liu2015investor,xu-cohen-2018-stock}.
Here, bullish (bearish) indicates the investor's expectation that the price of the mentioned stock will rise (decline).
When dealing with financial textual data, numbers have often been highlighted. Some studies attempt to extract numerical information~\cite{chen2019overview,chen2019crowdpt}. Others focus on the role of numbers in financial documents to explore reasoning skills~\cite{zhu-etal-2021-tat,chen-etal-2021-finqa,nan-etal-2022-fetaqa}. Further research suggests that the attention to numbers in financial documents can enhance downstream task performance, such as volatility forecasting~\cite{yang2022numhtml,shi-etal-2023-enhancing}. 
However, the relationship between financial sentiment analysis and numbers in text is rarely discussed. 
Building on this premise, this paper examines the efficacy of prompting LLMs to consider expert-designed hints, specifically the importance of numbers in understanding financial documents. 
Our findings indicate that the implementation of ``expert-designed hints,'' such as emphasizing the relevance of numerical data, substantially enhances the performance of sentiment analysis across various LLMs.

Financial sentiment analysis has been a longstanding topic in Natural Language Processing (NLP) research. Some studies focus on the writer's self-annotated labels, which are provided by the writer when posting tweets~\cite{li-shah-2017-learning,xing-etal-2020-financial}. Others make predictions about the readers’ sentiment~\cite{agic-etal-2010-towards,yuan-etal-2020-target,gaillat-etal-2018-ssix}. Additionally, there are studies that discuss the difference between writers' and readers' sentiment~\cite{maks-vossen-2013-sentiment,berengueres2017differences}. For example, \citeauthor{chen-etal-2020-issues}~(\citeyear{chen-etal-2020-issues}) show that writers' and readers' financial sentiments may differ based on a survey of 10K financial tweets. 
This discrepancy provides a basis for discussing perspective-taking within sentiment analysis, specifically, the extent to which models, acting as readers, can deduce a writer's sentiment, especially when it is implicitly conveyed. 
This paper aims to bridge this gap by utilizing the Fin-SoMe dataset~\cite{chen-etal-2020-issues}, comprising approximately 10,000 annotated social media posts from both the writer's and reader's viewpoints.
Our results suggest that the expert-designed hint activates perspective-taking and further enhances the performance of financial sentiment analysis.

In sum, this short paper aims to answer the following three research questions (RQ):

\vspace{2mm}

\noindent \textbf{RQ1}: Should LLMs be prompted to consider hints identified by experts, or are they inherently capable of recognizing such hints?

\vspace{2mm}

\noindent \textbf{RQ2}: To what extent can LLMs accurately apply perspective-taking ability when analyzing sentiments in financial social media data?

\vspace{2mm}

\noindent \textbf{RQ3}: Considering the importance of numerical data in financial documents, does the category of numbers affect sentiment analysis tasks?

\section{Related Work}
Theory of Mind (ToM)~\cite{baron1997mindblindness,baron1997parents,barnes2004perspective,baron2013understanding} has been a subject of interest for a long time and has recently regained attention. Discussions from the perspectives of vision~\cite{liu2022computational}, gaming~\cite{li-etal-2023-theory}, and psychology~\cite{van-dijk-etal-2023-theory} are provided, but discussions in sentiment analysis are scarce.
Perspective-taking involves understanding others' thoughts by empathizing with their viewpoints, a concept extensively explored within the realm of psychology under the ToM. LLMs have shown remarkable capabilities in grasping the semantics explicitly expressed in texts. Consequently, researchers in the NLP community have begun to investigate the applicability of ToM in evaluating whether models can empathize with human perspectives~\cite{liu2022computational,li-etal-2023-theory,van-dijk-etal-2023-theory,sclar-etal-2023-minding,sileo-lernould-2023-mindgames}. Despite numerous studies replicating psychological experiments to assess language models' abilities, investigations into perspective-taking for enhancing understanding of sentiments on social media are scant. 
 We propose a novel task design based on an existing dataset and discuss perspective-taking in financial sentiment analysis.

\section{Dataset}
We use the Fin-SoMe dataset \cite{chen-etal-2020-issues} in our experiments. A total of 10,000 tweets were collected from the social media platform Stocktwits,\footnote{https://stocktwits.com/} each labeled by the post's writer as either bullish or bearish. 
Reader sentiment was gauged by asking annotators to classify each tweet based on its content into bullish, bearish, or none categories. Tweets that did not explicitly convey sentiment were labeled as none. To discuss perspective-taking ability, instances that received a bullish or bearish label from the post's writer but were not labeled as bullish or bearish by the reader were deemed to require perspective-taking ability in this study. The dataset statistics is presented in Table~\ref{tab:Dataset Statistics}. This research does not involve training models; the entire dataset is treated as a test set to assess models' proficiency in financial sentiment analysis. The experimental results from this dataset inform the discussions of RQ1 and RQ2.

\section{Methods}
Our experiments aim to ascertain if LLMs inherently possess the ability to recognize and utilize the nuanced hint that experts deploy in deciphering financial documents.
Furthermore, we discuss the ability of LLMs to require perspective-taking ability.
We posit that if LLMs are inherently knowledgeable of and can apply these nuanced hints, their performance in sentiment analysis tasks should remain consistent, irrespective of whether these hints are explicitly highlighted within the prompt. Conversely, a notable discrepancy in performance would imply LLMs' lack of intrinsic capability to leverage these nuanced hints without explicit expert instruction.

\begin{table}[t]
  \centering
  \resizebox{\columnwidth}{!}{
    \begin{tabular}{l|rr}
     Writer Sentiment     & \multicolumn{1}{c}{Whole} & \multicolumn{1}{c}{Perspective-Taking} \\
    \hline
    Bullish &       8,573  &                        1,557  \\
    Bearish &       1,427  &                           277  \\
    \hline
    Total &     10,000  &                        1,834  \\
    \end{tabular}%
    }
  \caption{Dataset Statistics.}
  \label{tab:Dataset Statistics}%
\end{table}%

\begin{table*}[t]
  \centering
  \small
    \begin{tabular}{ll|rr|rr|rr}
    \multicolumn{1}{c}{\multirow{2}[3]{*}{LLM}} & \multicolumn{1}{c|}{\multirow{2}[3]{*}{Approach}} & \multicolumn{2}{c|}{Whole} & \multicolumn{2}{c|}{Perspective-Taking} & \multicolumn{2}{c}{Contains at Least One Number} \\
\cline{3-8}          &       & \multicolumn{1}{c}{Micro-F1} & \multicolumn{1}{c|}{Weighted-F1} & \multicolumn{1}{c}{Micro-F1} & \multicolumn{1}{c|}{Weighted-F1}  & \multicolumn{1}{c}{Micro-F1} & \multicolumn{1}{c}{Weighted-F1} \\
    \hline
    \multirow{3}[2]{*}{PaLM 2} & Simple Prompt & \textbf{80.84} & \textbf{84.49} & 70.77 & 76.96 & 80.61 & 84.29  \\
          & CoT   & 79.09 & 83.06 & 68.97 & 75.63 & 80.02 & 83.57 \\
          & CoT + Hint & 80.38 & 83.88 & \textbf{72.90}* & \textbf{78.30}*  &  \textbf{81.32}* & \textbf{84.39}*\\
    \hline
    \multirow{3}[2]{*}{Gemini Pro} & Simple Prompt & 68.26 & 77.48 & 42.80 & 55.71 & 69.06 & 78.01  \\
          & CoT   & 71.94 & 80.71 & 51.31 & 64.54 & 72.78 & 81.24 \\
          & CoT + Hint & \textbf{74.45}* & \textbf{81.66}* & \textbf{54.20}* & \textbf{65.91}* & \textbf{75.32}* & \textbf{82.18}* \\
    \hline
    \multirow{3}[2]{*}{GPT-3.5} & Simple Prompt & 68.43 & 77.43 & 41.44 & 54.88 & 69.23 & 77.95 \\
          & CoT   & 68.99 & 77.75 & 43.35 & 55.76 & 69.80 & 78.27 \\
          & CoT + Hint & \textbf{79.68}* & \textbf{83.72}* & \textbf{65.27}* & \textbf{73.22}* & \textbf{80.62}* & \textbf{84.23}*\\
    \hline
    \multirow{3}[1]{*}{GPT-4} & Simple Prompt & 78.01 & 84.59 & 54.80 & 67.68 & 78.93 & 85.13\\
          & CoT   & 73.35 & 82.04 & 48.64 & 62.57 & 74.21 & 82.58 \\
          & CoT + Hint & \textbf{81.86}* & \textbf{86.27}* & \textbf{62.00}* & \textbf{72.11}* & \textbf{82.82}* & \textbf{86.79}*\\
    \end{tabular}%
  \caption{Experimental results using the entire Fin-SoMe dataset and subsets within different settings. We use \textbf{bold font} to highlight the best performance among all approaches for the same LLM. An asterisk (*) indicates a significant difference when compared to the second-best performance across different approaches.}
  \label{tab:Experimental results}%
\end{table*}%

\begin{table*}[t]
  \centering
  \resizebox{\textwidth}{!}{
    \begin{tabular}{l|p{43em}c}
    Tweet (Bullish) & \multicolumn{2}{c}{\$MTBC haha whoever popped that fake ask above 2.60 earlier just loaded good 50k at 2.50 and 2.45!} \\
    \hline
          & \multicolumn{1}{c}{Analysis} & Sentiment \\
    \hline
    \multirow{2}[2]{*}{CoT}   & The user is commenting on a situation where someone apparently tried to manipulate the market price of \$MTBC but ended up buying at a lower price. The user seems to be mocking those who fell for the trick. & \multirow{2}[2]{*}{Neutral} \\
    \hline
    \multirow{3}[4]{*}{CoT + Hint} & The sentiment of this tweet appears to be bullish for the person who managed to buy at lower prices, as they seem to have manipulated the market in their favor. ... Overall, considering the tone of mockery and satisfaction in the tweet, it leans more towards bullish sentiment as it implies an expectation of profit from the lower purchase prices. & \multirow{3}[4]{*}{Bullish} \\
    \end{tabular}%
    }
  \caption{Case Study.}
  \label{tab:Case Study}%
\end{table*}%

Furthermore, our methodology incorporates not only the direct sentiment prediction approach, termed as the Simple Prompt, but also includes the Chain-of-Thought (CoT) technique~\cite{wei2022chain}, which involves eliciting an analytical reasoning before concluding with a prediction. To evaluate the potential enhancement in performance through the integration of nuanced hints, we augment our prompts with a directive emphasizing the importance of numerical data or statistics embedded within the tweets, as these elements could be pivotal in deducing the overarching sentiment. This approach is inspired by numerous studies on number comprehension and reasoning within financial texts~\cite{zhu-etal-2021-tat,chen-etal-2021-finqa,nan-etal-2022-fetaqa,yang2022numhtml,shi-etal-2023-enhancing}, which collectively suggest that numerical information plays a critical role in understanding intentions or making informed decisions. In short, we add a sentence, \textit{``Focus particularly on any numerical data or statistics present in the tweet, as these figures may be crucial in determining the overall sentiment.''}, in the CoT prompt as a hint in the experiment. 

Our analysis encompasses four LLMs, including PaLM 2~\cite{anil2023palm}, Gemini Pro,\footnote{https://deepmind.google/technologies/gemini/} GPT-3.5, and GPT-4.\footnote{https://platform.openai.com/docs/models} To quantify their performance, we employ both Micro-F1 and Weighted-F1 scores. Furthermore, we utilize McNemar's test~\cite{mcnemar1947note} to ascertain the statistical significance of performance disparities among the models, setting the significance threshold at $\alpha = 0.05$.

\section{Experimental Results}
\subsection{Overall Performance}
Table~\ref{tab:Experimental results} shows the experimental results. First, when employing a Simple Prompt, the performance of PaLM 2 is superior to all other LLMs. Second, the CoT approach does not invariably result in enhanced performance; it improves performance in two out of four LLMs (Gemini Pro and GPT-3.5). Third, the inclusion of an expert-designed hint consistently improves performance across all LLMs compared to using CoT. Moreover, it significantly outperforms other methods in three out of four LLMs. Although it marginally underperforms compared to the Simple Prompt with PaLM 2, the difference is not statistically significant. This finding addresses RQ1 and rejects our hypothesis. The experimental outcomes suggest that hints, specifically numbers, commonly utilized in prior research for analyzing financial documents, are not automatically leveraged by LLMs in financial text analysis without explicit guidance. A concise summary of these experiments and findings indicates that a simple yet crucial hint provided by experts can significantly enhance the results of sentiment analysis in financial social media data.
In conclusion, our primary objective was to investigate whether the expert-designed hint utilized by experts affects the performance of LLMs in financial sentiment analysis. Our findings affirmatively answer RQ1, indicating that the addition of an expert-designed hint generally improves performance.

\subsection{Perspective-Taking Subset}
To address RQ2, we focus on analyzing the perspective-taking subset. The experimental results in Table~\ref{tab:Experimental results} initially reveal that this subset poses more challenges, as evidenced by lower performance compared to experiments involving the entire dataset, regardless of the LLMs and approaches applied. Additionally, the expert-designed hint consistently yields superior performance in this subset across all LLMs. These results suggest that the expert-designed hint facilitates the activation of the perspective-taking capability of LLMs. Table~\ref{tab:Case Study} presents an example tweet labeled as bullish by the writer of the tweet. The analysis and sentiment labels generated by GPT-4 using different approaches are also provided. This example firstly highlights the significance of numerical information in financial texts. Ignoring the four numbers within this 18-token tweet would result in a loss of considerable information. Secondly, it demonstrates that GPT-4 can comprehend the content and deliver a precise analysis. Nonetheless, the analysis and sentiment label change upon the addition of a hint, leading to a more thorough analysis by recognizing the tone and emotions of the writer and correctly aligning the sentiment label with that of the writer.
To sum up, our further analysis reveals that part of the improvement is attributable to the activation of perspective-taking, thereby answering RQ2.

\subsection{Tweets with Numbers}
Given that the expert-designed hint pertains to numbers, it is crucial to verify whether this hint indeed enhances performance in tweets containing numbers. To evaluate the outcomes, we assess the performance of all methods on a subset where each tweet includes at least one number. The results are presented in Table~\ref{tab:Experimental results}. Generally, the performance on this subset surpasses that observed with the entire dataset. 
However, the highest performances with CoT + Hint, regardless of the LLM employed, exceeded those achieved with the full dataset. This suggests that the hint significantly aids in financial sentiment analysis. Furthermore, these results reveal a marked distinction between different methodologies when utilizing PaLM 2, indicating that, although the performance differential between the Simple Prompt and CoT + Hint approaches is negligible for the entire dataset, the performance gain within the subset containing numbers is significant. In summary, these findings demonstrate that the expert-designed hint substantially facilitates financial sentiment analysis by merely reminding models of its presence.

\subsection{Category of the Number}
Given the significance of numbers in social media data, we further analyze the enhancement in tweets containing various types of numbers.
To discuss RQ3, we further compare the FinSoMe dataset with the FinNum dataset~\cite{chen2019overview}. The FinNum dataset comprises annotations for 8,868 numbers found in financial social media posts, categorized into seven types specifically designed for interpreting numbers in financial contexts: monetary, percentage, option, indicator, temporal, quantity, and product/version. We identified 6,493 tweets present in both the FinSoMe and FinNum datasets and used this subset to explore how differences in categories of numbers influence sentiment analysis performance.

We divided this set into seven groups based on the number of category labels in the FinNum dataset. The improvement is calculated based on the micro-F1 performance using Simple Prompt and CoT + Hint. Table~\ref{tab:Improvement} presents the results. Firstly, 37.53\% of instances contain numbers related to the Monetary category, and we observed that the enhancement in this category is notably high compared to other groups. Additionally, significant improvement is a phenomenon observed regardless of the LLMs applied. The example in Table~\ref{tab:Case Study} also shows the importance of numbers in the Monetary category, with all numbers (2.50, 50k, 2.50, and 2.45) being related to monetary values. Secondly, enhancement is observed in most groups, with only a few cases showing worse performance. This indicates that a simple hint can effectively guide LLMs to perform more comprehensive sentiment analysis, focusing on aspects considered important by experts.

\begin{table}[t]
  \centering
    \resizebox{\columnwidth}{!}{
    \begin{tabular}{l|r|r|r|r|r}
    \multicolumn{1}{c|}{Category} & \multicolumn{1}{c|}{Instance (\%)} & \multicolumn{1}{c|}{PaLM 2} & \multicolumn{1}{c|}{Gemini Pro} & \multicolumn{1}{c|}{GPT-3.5} & \multicolumn{1}{c}{GPT-4} \\
    \hline
    Monetary & 37.53 & 11.98 & 42.35 & 59.37 & 22.64 \\
    \hline
    Temporal & 30.23 & 0.80 & 18.16 & 32.85 & 7.74 \\ 
    \hline
    Percent & 13.32 & -1.38 & 5.72 & 11.16 & 7.79 \\
    \hline
    Quantity & 12.46 & 1.48 & 9.77 & 14.83 & 4.20 \\
    \hline
    Indicator & 2.43 & 0.00 & 0.00 & 10.76 & 4.43 \\
    \hline
    Option & 2.28 & -8.98 & 8.98 & 25.17 & 2.99 \\
    \hline
    Product Number & 1.74 & 1.77 & 10.62 & 17.70 & 13.27 \\
    \end{tabular}%
    }
  \caption{Improvement (\%) in the subset of tweets containing a number in the target category.}
  \label{tab:Improvement}%
\end{table}%

\section{Conclusion}
This study investigates the impact of employing expert-designed hint on the performance of LLMs in financial sentiment analysis. We find that LLMs do not inherently utilize subtle hints crucial for sentiment analysis without explicit instruction. Introducing a simple, expert-derived hint that highlights the significance of numerical data substantially improves the models' capability to identify sentiments. This enhancement is especially notable in scenarios requiring perspective-taking, where the models must deduce the sentiment implied by the writer, highlighting the necessity of explicit guidance in financial sentiment analysis tasks.

\section*{Limitation}
The limitations of this study are discussed as follows.

Firstly, the findings of this study are primarily based on financial social media data, particularly from the Stocktwits platform. This focus may limit the generalizability of our conclusions to other domains or types of social media content. Future studies could explore whether the observed benefits of perspective-taking and expert-designed hint extend to other domains, such as healthcare or politics, where sentiment analysis is equally critical.

Secondly, this study simplifies the concept of perspective-taking. However, perspective-taking in human communication is a complex, multi-dimensional process that involves understanding emotional states, intentions, and contextual factors. Future work could aim to model these additional layers of complexity to achieve a more holistic understanding of sentiment in social media texts.

Another limitation is the focus on only four LLMs in our experiments. While these models are among the most advanced at the time of our study, the rapidly evolving field of natural language processing continually introduces new models that may offer different insights into the challenges of financial sentiment analysis. Testing our approach with a wider array of LLMs could provide a more comprehensive understanding of its effectiveness.

Lastly, our study's focus on numerical data as a key element of financial sentiment analysis may overlook other important factors that influence sentiment interpretation, such as linguistic subtleties, cultural references, or domain-specific knowledge. Incorporating these dimensions into future research could provide a more holistic understanding of the challenges and opportunities in applying large language models to financial sentiment analysis.

\bibliography{anthology}
\bibliographystyle{acl_natbib}

\end{document}